%% file: main.tex
\documentclass[11pt]{article}
\usepackage[a4paper,margin=1in]{geometry}
\usepackage[T1]{fontenc}
\usepackage[utf8]{inputenc}
\usepackage{amsmath,amssymb,graphicx,hyperref}
\usepackage{multirow,booktabs}
\usepackage{makecell}
\usepackage{pgfplots}
\pgfplotsset{compat = newest}
\usepackage{tikz}
\usetikzlibrary{matrix,positioning,external}
\hypersetup{
    colorlinks=true,
    linkcolor=black,
    citecolor=bluecite,
    urlcolor=bluecite
}

\title{Efficient Few-Shot Learning for Edge AI \\via Knowledge Distillation on MobileViT}
\author{
Shuhei Tsuyuki$^{1,2}$ \and
Reda Bensaid$^{3}$ \and
J\'er\'emy Morlier$^{3}$ \and
Mathieu L\'eonardon$^{3}$ \and
Naoya Onizawa$^{1}$ \and
Vincent Gripon$^{3}$ \and
Takahiro Hanyu$^{1}$
\\[0.5em]
\normalsize $^{1}$Research Institute of Electrical Communication, Tohoku University, Japan\\
\normalsize $^{2}$Graduate School of Engineering, Tohoku University, Japan\\
\normalsize $^{3}$IMT Atlantique, Lab-STICC, UMR CNRS 6285, F-29238 Brest, France
}
\date{}

\input{colors}

\begin{document}
\maketitle

\begin{abstract}
Efficient and adaptable deep learning models are an important area of deep learning research, driven by the need for highly efficient models on edge devices. Few-shot learning enables the use of deep learning models in low-data regimes, a capability that is highly sought after in real-world applications where collecting large annotated datasets is costly or impractical. This challenge is particularly relevant in edge scenarios, where connectivity may be limited, low-latency responses are required, or energy consumption constraints are critical. We propose and evaluate a pre-training method for the MobileViT backbone designed for edge computing. Specifically, we employ knowledge distillation, which transfers the generalization ability of a large-scale teacher model to a lightweight student model. This method achieves accuracy improvements of 14\% and 6.7\% for one-shot and five-shot classification, respectively, on the MiniImageNet benchmark, compared to the ResNet12 baseline, while reducing by 69\% the number of parameters and by 88\% the computational complexity of the model, in FLOPs. Furthermore, we deployed the proposed models on a Jetson Orin Nano platform and measured power consumption directly at the power supply, showing that the dynamic energy consumption is reduced by 37\% with a latency of 2.6 ms. These results demonstrate that the proposed method is a promising and practical solution for deploying few-shot learning models on edge AI hardware.

\end{abstract}

\noindent\textbf{Keywords:} Few-shot learning, Edge AI, MobileViT, Knowledge distillation

\section{Introduction}
The rapid growth of artificial intelligence (AI) has led to dramatic increases in the computational complexity, power consumption, and dataset size required for training. While these demands can be met using large-scale computing infrastructure, they still pose a significant burden for edge AI devices, which must operate under severe resource constraints. Therefore, approaches that reduce training costs and data dependency while maintaining accuracy are essential. Few-shot learning (FSL) offers a promising solution, enabling generalization from a limited number of samples while reducing computational and data costs.

FSL has been widely studied in fields such as natural language processing, image recognition, and image classification. Existing research, such as Matching Networks~\cite{vinyals2016matching}, Prototypical Networks~\cite{snell2017prototypical}, and EASY~\cite{bendou2022easy}, which is used as the baseline in this work, has shown that FSL can achieve competitive performance. Furthermore, research such as PEFSL~\cite{grativol2024pefsl} and EdgeFSL~\cite{kanda2024design} has demonstrated the deployment of FSL on edge devices, highlighting the computational benefits of FSL for resource-constrained applications.

However, challenges remain in leveraging FSL at the edge. Most prior work focuses on ResNet-based backbones, limiting exploration of more efficient alternative architectures. Additionally, due to the severe resource constraints of edge devices, accuracy often remains insufficient compared to large-scale models~\cite{xu2024towards}. Therefore, a method for achieving higher accuracy on lightweight backbones is needed. 

In this work, we evaluate multiple efficient backbones for FSL and adopt MobileViT~\cite{mehta2021mobilevit}, a hybrid CNN+Vision Transformer architecture that balances local feature extraction with global context modeling. To further improve performance, we pre-train MobileViT using knowledge distillation~\cite{hinton2015distillation}, transferring the generalization ability of a large teacher model to a lightweight student. Moreover, we demonstrate the practical deployment of the proposed models on a Jetson Orin Nano, measuring power consumption directly at the power supply and showing a 37\% reduction in dynamic energy usage compared to a ResNet baseline.

Overall, our contributions include the development of a lightweight yet accurate FSL backbone, the use of knowledge distillation for improved generalization, and the demonstration of energy-efficient deployment on real edge hardware.

\begin{figure}[t]
    \centering
    \includegraphics[width=0.9\linewidth]{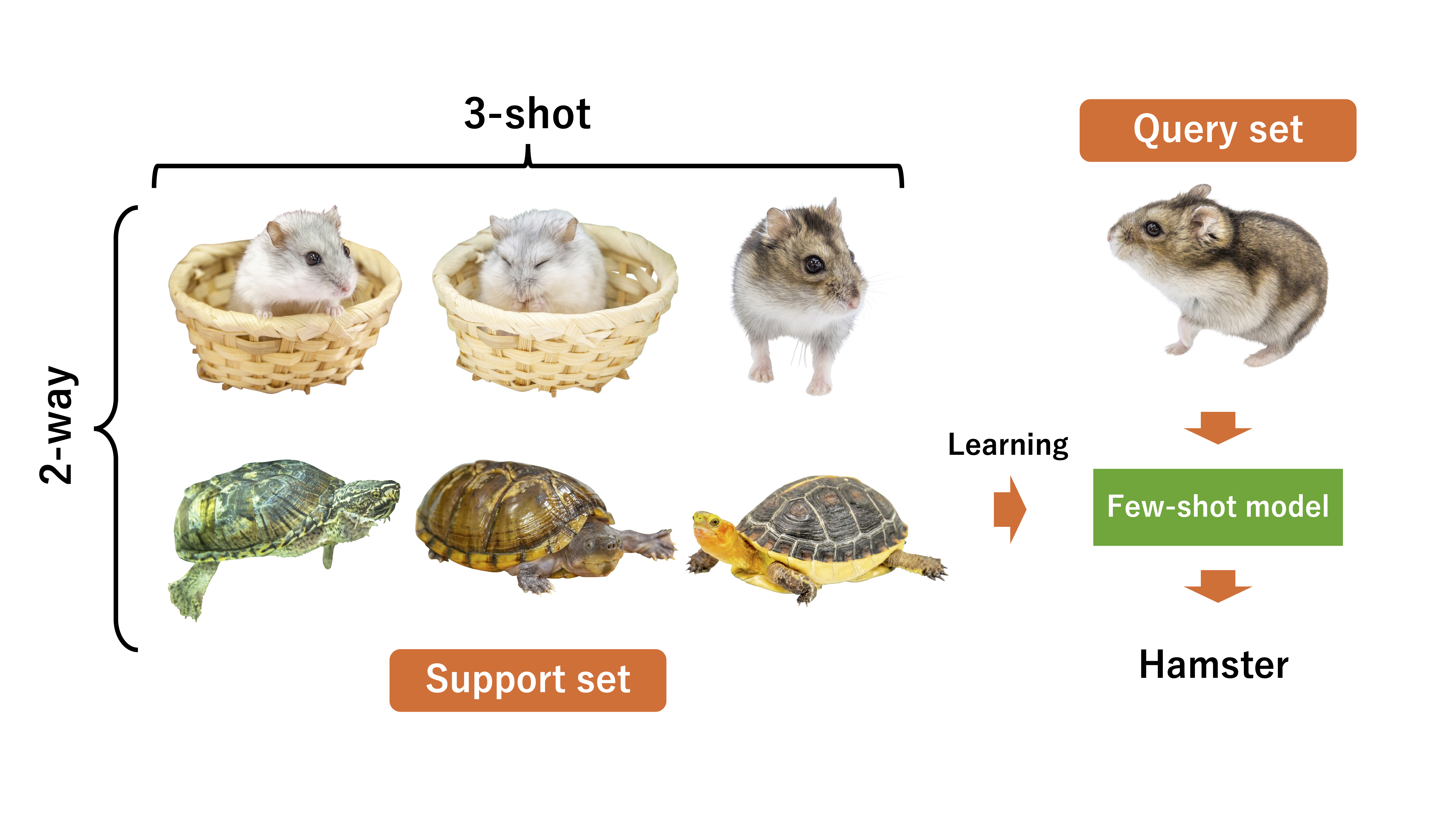}
    \caption{Few-Shot Learning uses a small support set to train and a query set to test. The number of classes is called the way, and the number of examples per class is the shots.}
    \label{fig:fsl}
\end{figure}
\section{Background and Related Work}

\subsection{Few-Shot Learning}

\begin{figure}[t]
    \centering
    \includegraphics[width=0.9\linewidth]{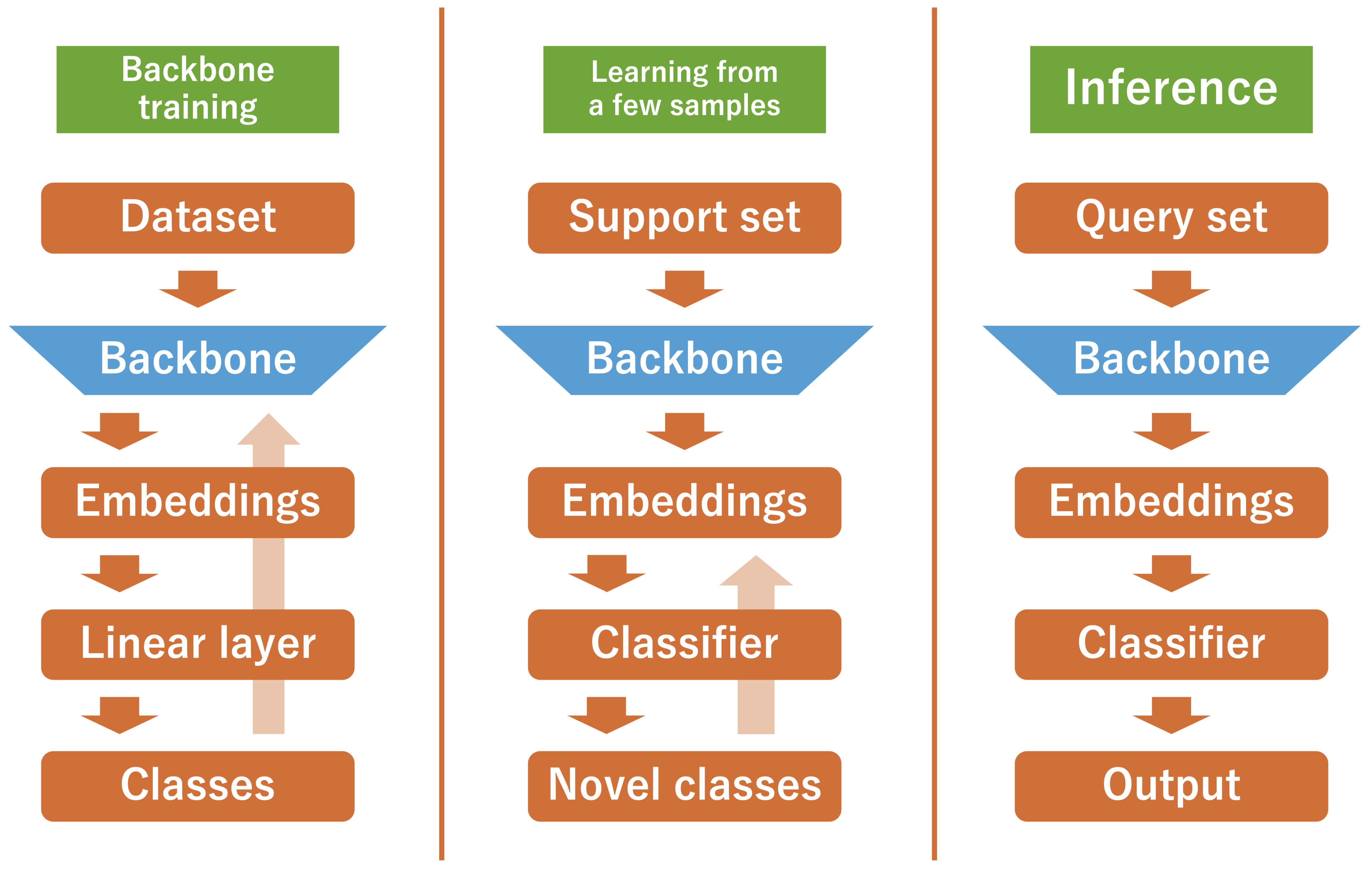}
    \caption{FSL consists of three steps: backbone learning for feature extraction, learning from a small amount of data using a pre-trained backbone, and evaluating the inference ability for classification tasks.}
    \label{fig:easy}
\end{figure}

Few-shot learning (FSL) is a technique that enables learning from a small number of labeled examples, as shown in Figure~\ref{fig:fsl}. It is particularly effective in scenarios where data availability is limited, such as medical image diagnosis of rare diseases. FSL uses two types of data: a \textit{support set} and a \textit{query set}. The support set provides a small number of labeled examples to guide learning, and the query set contains examples from the same classes used to evaluate the model's inference ability. As an example, the overall flow of FSL in EASY~\cite{bendou2022easy}, the baseline of this study, is shown in Figure~\ref{fig:easy}.

\medskip\noindent\textbf{Backbone training.}
The backbone (feature extractor) is first trained on a large dataset that excludes the images and classes used in the support set. After pre-training, the parameters of the backbone are fixed, as in EASY~\cite{bendou2022easy} and in our study. However, not all FSL methods fix the backbone.

\medskip\noindent\textbf{Learning from a few samples.}
Features are extracted from the support set images using the backbone, and a classifier---such as the Nearest Class Mean (NCM) method described below---is trained on these features.

\medskip\noindent\textbf{Inference.}
During inference, features are extracted from the query set images using the trained backbone and then passed to the classifier to produce classification results.

\subsection{Nearest Class Mean}

The Nearest Class Mean (NCM)~\cite{snell2017prototypical} classifier represents each class by the mean of its support feature embeddings, called a \textit{prototype}. During inference, a query sample is embedded by the backbone network and assigned to the class whose prototype is closest in Euclidean distance.

Given the support set $S_i$ of class $i$, the prototype is computed as
\begin{equation}
    \mathbf{c}_i = \frac{1}{|S_i|} \sum_{\mathbf{z} \in S_i} \mathbf{z}.
\end{equation}
Classification is then performed by selecting the nearest prototype in feature space.
Due to its simplicity and low computational cost, NCM is well suited for few-shot learning and resource-constrained edge AI applications.

\subsection{EASY}

As mentioned in the introduction, EASY~\cite{bendou2022easy} was proposed as a simple yet strong baseline for FSL. EASY enhances a ResNet12~\cite{he2016resnet} backbone through supervised classification, auxiliary rotation prediction, and data augmentation such as manifold mixup~\cite{verma2019manifold}. 
In the classification stage, it refines class prototypes by applying feature normalization and centering, and further improves stability using ensemble features and augmented shots. EASY adopts the NCM classifier in the inductive setting and soft k-means in the transductive setting. 
Despite its simplicity, EASY achieves competitive performance on standard benchmarks such as MiniImageNet and TieredImageNet, establishing it as a widely adopted baseline in FSL.

\subsection{Knowledge Distillation}

Recent few-shot learning methods increasingly rely on large foundation models to adapt to tasks with limited data~\cite{xu2024towards}. However, their high computational and energy costs limit deployment in edge AI, motivating the use of lightweight vision backbones~\cite{sharshar2025vision}. Knowledge distillation (KD)~\cite{hinton2015distillation} provides a practical solution by transferring knowledge from a large teacher to a compact student while preserving efficient inference.

In few-shot learning, prior work has explored specialized KD formulations and frameworks~\cite{smkd2023, wecolora2024, xu2024towards}, often introducing complex objectives or additional training overhead. In contrast, our work focuses on deployable edge-oriented learners, prioritizing simplicity, inference efficiency, and energy-aware performance rather than extending KD formulations.

\section{Proposed Method}

\begin{figure}[t]
    \centering
    \includegraphics[width=0.9\linewidth]{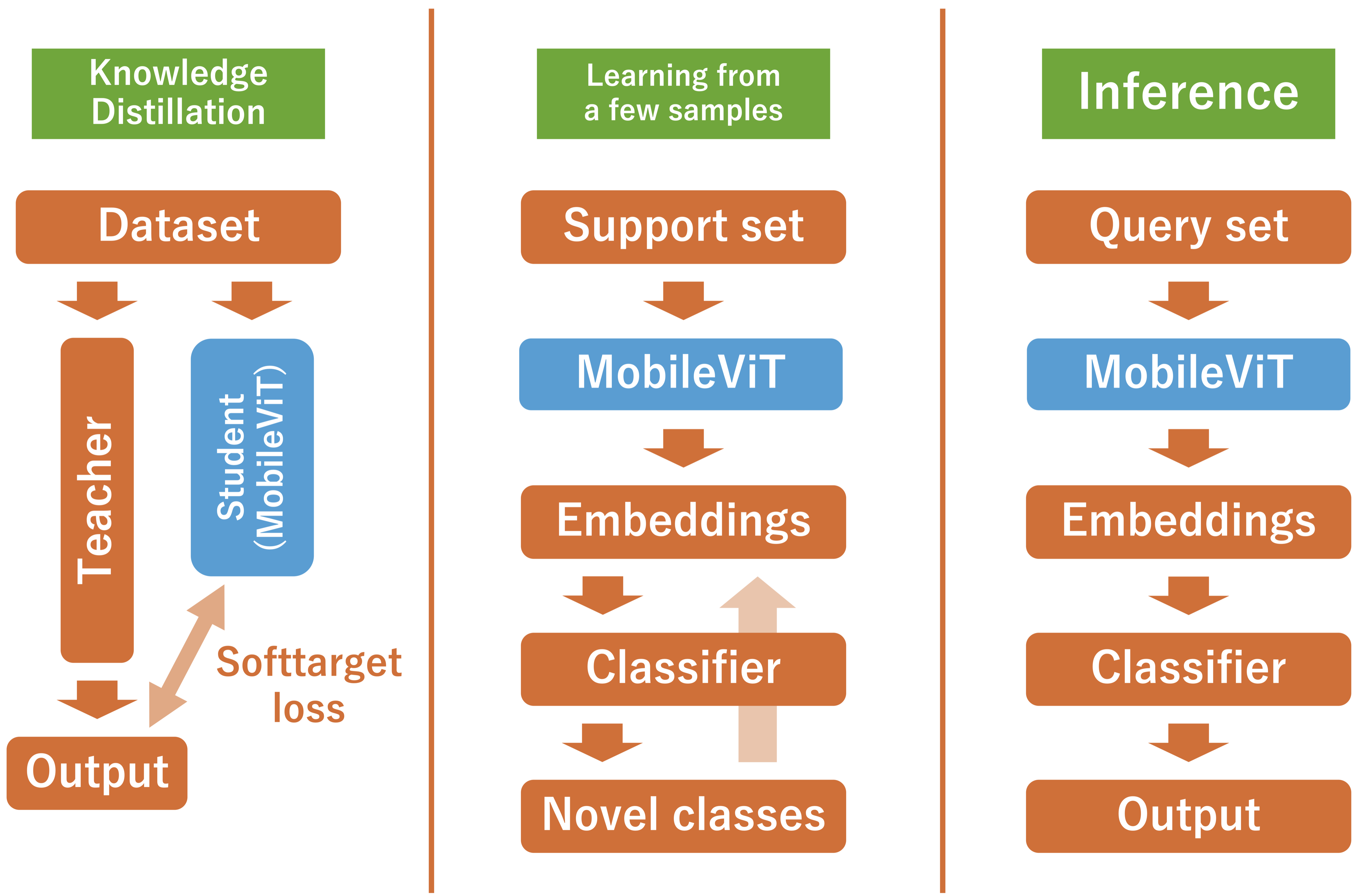}
    \caption{Overview of the proposed method: MobileViT-XXS is used as the backbone, and knowledge distillation is performed as pre-training. After this, the backbone is not updated, and training and inference ability are evaluated using a small number of samples, as in Figure~\ref{fig:easy}.}
    \label{fig:overview}
\end{figure}

An overview of the method proposed in this study is shown in Figure~\ref{fig:overview}. In the baseline, ResNet12~\cite{he2016resnet} was pre-trained using the EASY~\cite{bendou2022easy} method, whereas in the proposed method, MobileViT~\cite{mehta2021mobilevit} is used as the backbone and pre-training is performed using knowledge distillation~\cite{hinton2015distillation}. After pre-training, the support set is used to adapt to few-shot learning.

\subsection{Lightweight Backbones}

\begin{figure}[t]
    \centering
    \input{figure/backbones}
    \vspace{-10pt}
    \caption{Performance comparison of EfficientNet (B0–B3), MobileNetV3 (S, L), and MobileViT (S, XS, XXS) backbones in the first stage of the EASY pipeline without ensemble features or augmented shots. All models were trained on $84 \times 84$ images with batch size 376 for 100 epochs using a learning rate of 0.01 with $\gamma=0.1$. Accuracy after pre-training is reported as a proxy for backbone quality in few-shot learning.}
    \label{fig:backbones}
\end{figure}
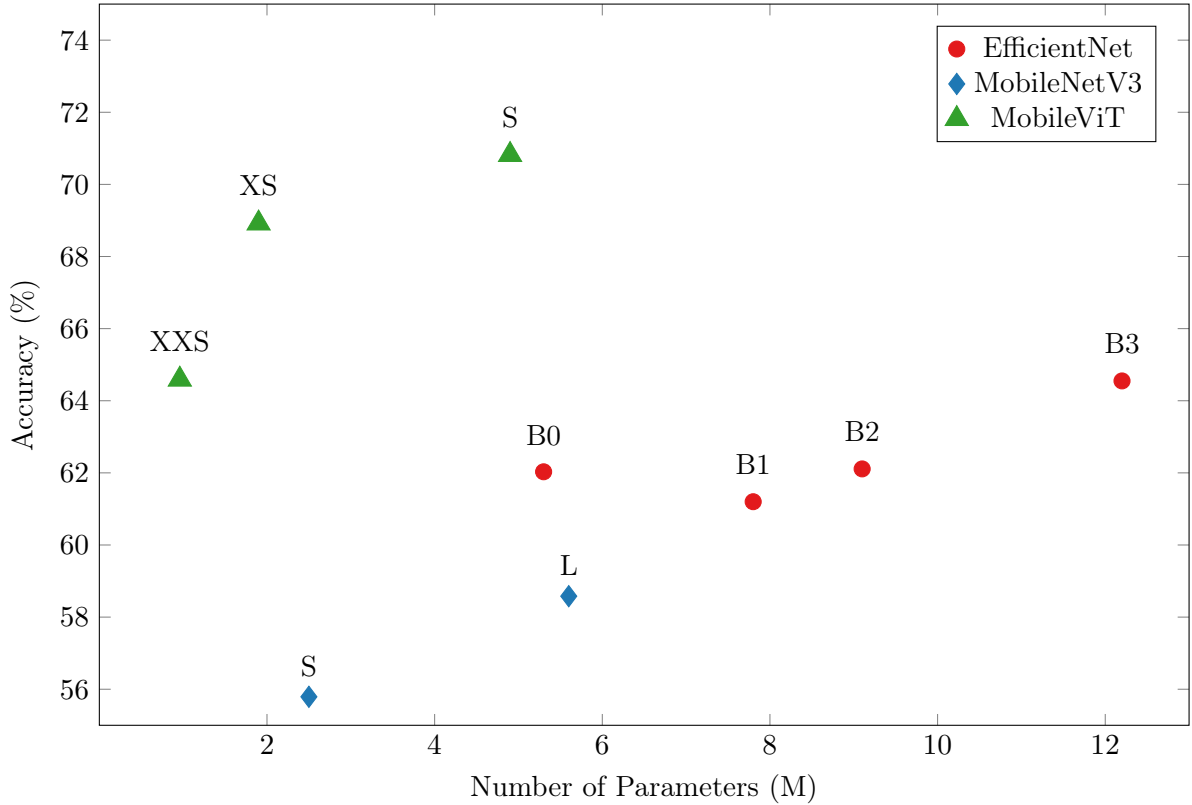

Lightweight neural network backbones are essential for edge AI, where resource constraints impose strict limits on model size and complexity. Among the most widely adopted families, EfficientNet~\cite{tan2019efficientnet} introduced a compound scaling approach to balance accuracy and efficiency across multiple model sizes, while MobileNetV3~\cite{howard2019searching} combined depthwise separable convolutions with neural architecture search and squeeze-and-excitation blocks to achieve high performance at low cost. 
MobileViT~\cite{mehta2021mobilevit} takes a hybrid approach, integrating local convolutions with global self-attention inspired by Vision Transformers (ViT)~\cite{dosovitskiy2021image}, providing both efficiency and strong representational power.

To compare these families under consistent conditions, we pre-trained representatives of EfficientNet, MobileNetV3, and MobileViT within the first stage of the EASY~\cite{bendou2022easy} pipeline, without ensemble features or augmented shots (E/AS). Figure~\ref{fig:backbones} summarizes their performance, serving as a proxy for backbone quality in other scenarios, as validated in EASY~\cite{bendou2022easy}.
MobileViT consistently outperforms MobileNetV3 and EfficientNet across all configurations. We attribute this to its hybrid design, which combines the local feature extraction ability of CNNs with the global self-attention mechanisms of Vision Transformers (ViTs). In few-shot learning, where only a limited number of samples per class are available, capturing global relationships between spatial regions becomes critical for generalization, as the model cannot rely solely on learning all possible local variations. Moreover, ViTs and hybrid architectures like MobileViT are known to produce more domain-agnostic representations than purely convolutional networks, improving transferability to novel classes. These properties explain why MobileViT achieves higher accuracy in this FSL setting despite its lightweight design.

\subsection{Knowledge Distillation}

In this work, we adopt ViT-S/14 from the DINOv2~\cite{dino2024} family as the teacher model, which provides strong general-purpose visual representations. To ensure robustness and reproducibility, we employ a simple distillation strategy: instead of soft/hard target weighting or temperature scaling, the student is trained to align its feature embeddings with those of the teacher using a mean squared error (MSE) loss. This design avoids additional hyperparameters and emphasizes practical and stable knowledge transfer for edge few-shot learning.

\section{Evaluation}

This section presents experimental results demonstrating the effectiveness of the proposed method. Unless otherwise noted, performance evaluation is based on the inference capability of a five-way system, and when evaluating FSL inference, the number of samples for each class is set to 15. To ensure stable performance estimation, 10,000 inferences are performed for all of these samples, and this is performed five times independently with different random seeds. The experimental results include the average accuracy as well as a 95\% confidence interval calculated using a normal distribution approximation.

\subsection{Comparison with Baselines and Hyperparameter Tuning}
\label{subsec:comparison_baselines}

Table~\ref{tab:comparison_baselines} summarizes the performance of EASY~\cite{bendou2022easy} and the proposed method. The dataset used for evaluation was MiniImageNet, and inference capabilities were evaluated in both one-shot and five-shot runs. MiniImageNet, consisting of 100 classes with 600 images each, is a widely used benchmark for evaluating few-shot learning. 

To ensure a fair comparison, we also trained MobileViT-XXS~\cite{mehta2021mobilevit} using the original EASY~\cite{bendou2022easy} method. This allows us to isolate the benefit brought by knowledge distillation compared to standard pre-training. Table~\ref{tab:params_flops} further reports the number of parameters and FLOPs (Floating-Point Operations) for ResNet12~\cite{he2016resnet}, used in the original EASY, and MobileViT-XXS~\cite{mehta2021mobilevit}, used in our method.

During our experiments, we varied both the learning rate and the number of epochs for knowledge distillation~\cite{hinton2015distillation} to identify the most effective settings. While our knowledge distillation approach significantly improves performance, we observed that the learning rate is an important factor in achieving optimal results. Careful tuning of the learning rate ensures stable training and allows the student model to effectively inherit knowledge from the teacher. In our experiments, a learning rate of 0.001 consistently provided the best trade-off between convergence speed and accuracy. We trained the models for 100 epochs.

Overall, the results in Table~\ref{tab:comparison_baselines} and Table~\ref{tab:params_flops} show that the proposed method not only reduces computational complexity and model size but also significantly improves accuracy in both the 1-shot and 5-shot settings compared to all baselines.

\begin{table}[t]
\centering
\caption{Few-shot classification accuracy on MiniImageNet for EASY (ResNet12), EASY + MobileViT-XXS, and KD + MobileViT-XXS.}
\vspace{2mm}
\label{tab:comparison_baselines}
\begin{tabular}{lcc}
\hline
\makecell{Method} & 1-shot & 5-shot \\
\hline
\makecell{EASY \\ + ResNet12} & 70.84$\pm$0.19 & 86.28$\pm$0.12 \\
\makecell{EASY \\ + MobileViT-XXS (ours)} & 75.08$\pm$0.10 & 88.43$\pm$0.08 \\
\makecell{EASY + KD \\ + MobileViT-XXS (ours)} & \textbf{85.08$\pm$0.06} & \textbf{92.96$\pm$0.05} \\
\hline
\end{tabular}
\end{table}

\begin{table}[t]
\centering
\caption{Comparison of the number of parameters and FLOPs of ResNet12 and MobileViT-XXS.}
\vspace{2mm}
\label{tab:params_flops}
\begin{tabular}{lcc}
\hline
Backbone & Param & FLOPs \\
\hline
ResNet12 & 3.109M & 4.405 G \\
MobileViT-XXS & 0.961M & 0.512 G \\
\hline
\end{tabular}
\end{table}

\subsection{Evaluation on Additional Datasets}
\label{subsec:additional_datasets}

Finally, to confirm the generalization ability of MobileViT-XXS trained using the proposed method, we evaluated it on additional benchmark datasets. For comparison, we also evaluated MobileViT-XXS trained using the EASY method without knowledge distillation. 
As shown in Tables~\ref{tab:additional_datasets_1shot} and~\ref{tab:additional_datasets_5shot}, our KD + EASY approach consistently outperforms the baseline EASY-trained MobileViT-XXS in most 1-shot and 5-shot scenarios. Notably, large improvements are observed on datasets such as MiniImageNet, ImageNet, and Caltech101 in the 1-shot setting, where accuracy gains range from 4\% to 10\%. These results demonstrate that knowledge distillation effectively enhances the generalization capability of the lightweight backbone, yielding higher accuracy across diverse datasets while maintaining the efficiency advantages of MobileViT-XXS. Only in a few cases (e.g., EuroSAT, Flowers102, Caltech101 in 5-shot) the baseline slightly surpasses KD + EASY, highlighting that the method is broadly robust across different domains.
\begin{table}[t]
\centering
\caption{1-shot few-shot classification accuracy on various datasets for EASY + MobileViT-XXS and KD + MobileViT-XXS.}
\vspace{2mm}
\label{tab:additional_datasets_1shot}
\begin{tabular}{lcc}
\hline
Dataset & \makecell{EASY + \\MobileViT-XXS} & \makecell{EASY + KD + \\MobileViT-XXS} \\
\hline
MiniImageNet & 75.08$\pm$0.10 & \textbf{85.08$\pm$0.06} \\
ImageNet & 75.05$\pm$0.13 & \textbf{82.61$\pm$0.08} \\
Caltech101 & 74.59$\pm$0.10 & \textbf{79.03$\pm$0.09} \\
OxfordPets & 71.04$\pm$0.07 & \textbf{84.94$\pm$0.14} \\
Flowers102 & 68.61$\pm$0.05 & \textbf{70.82$\pm$0.14} \\
Food101 & 39.18$\pm$0.07 & \textbf{47.13$\pm$0.10} \\
FGVCAircraft & 31.36$\pm$0.03 & \textbf{33.99$\pm$0.12} \\
EuroSAT & \textbf{60.36$\pm$0.11} & 58.77$\pm$0.06 \\
ImageNet-A & 39.58$\pm$0.14 & \textbf{41.23$\pm$0.10} \\
\hline
\end{tabular}
\end{table}

\begin{table}[t]
\centering
\caption{5-shot few-shot classification accuracy on various datasets for EASY + MobileViT-XXS and KD + MobileViT-XXS.}
\vspace{2mm}
\label{tab:additional_datasets_5shot}
\begin{tabular}{lcc}
\hline
Dataset & \makecell{EASY + \\MobileViT-XXS} & \makecell{EASY + KD + \\ MobileViT-XXS} \\
\hline
MiniImageNet & 88.43$\pm$0.08 & \textbf{92.96$\pm$0.05} \\
ImageNet & 87.70$\pm$0.04 & \textbf{90.80$\pm$0.07} \\
Caltech101 & \textbf{89.09$\pm$0.08} & 88.88$\pm$0.05 \\
OxfordPets & 90.19$\pm$0.10 & \textbf{94.07$\pm$0.06} \\
Flowers102 & \textbf{87.97$\pm$0.06} & 85.14$\pm$0.07 \\
Food101 & 60.49$\pm$0.09 & \textbf{62.77$\pm$0.09} \\
FGVCAircraft & 38.04$\pm$0.09 & \textbf{42.08$\pm$0.09} \\
EuroSAT & \textbf{74.79$\pm$0.08} & 72.91$\pm$0.03 \\
ImageNet-A & 50.82$\pm$0.06 & \textbf{53.22$\pm$0.07} \\
\hline
\end{tabular}
\end{table}

\section{Energy Consumption Analysis on Jetson Orin Nano}

\subsection{Energy Measurement Setup}

Energy consumption was measured on the \textit{Monolithe}\footnote{https://monolithe.proj.lip6.fr/} platform~\cite{cassagne2025dalek}, a cluster of single-board computers (SBCs) designed to evaluate the suitability of SoC-based architectures for embedded applications. Each SBC is powered through a dedicated high-frequency (5 kHz) measurement board, enabling real-time monitoring of voltage, current, and power consumption directly at the power supply input rather than relying on software tools such as \texttt{tegrastats}.

\subsection{Experimental Results}

\begin{table}[ht]
\centering
\caption{Energy and performance metrics on Jetson Orin Nano at 15 W power mode (transposed).}
\vspace{2mm}
\label{tab:energy}
\begin{tabular}{lccc}
\hline
\textbf{Metric} & \textbf{Idle} & \textbf{ResNet} & \textbf{MobileViT-XXS} \\
\hline
Power (W)       & 4.0   & 19.9 & 14.0 \\
Latency (ms)    & --    & 3.6  & 2.6  \\
Throughput (ips)& --    & 280  & 392  \\
Energy/inf (J)  & --    & 0.072 & 0.036 \\
\hline
\end{tabular}
\end{table}
We measured the power consumption, latency, throughput, and energy per inference for ResNet and MobileViT-XXS on a Jetson Orin Nano set to a 15 W power mode. Results are summarized in Table~\ref{tab:energy}. Energy per inference is computed as $E = P \times \frac{\text{Latency}}{1000}$.

The results highlight several important points. First, although ResNet requires fewer FLOPs, it exhibits higher latency than MobileViT-XXS. This indicates that FLOPs alone are not a reliable predictor of inference speed, likely due to memory access patterns and parallelization differences between CNN and Transformer architectures. Second, MobileViT-XXS consumes less power overall (14 W vs 19.9 W) and has lower energy per inference. By subtracting the idle power (4 W) from the measured power, we compute the dynamic power, which is reduced by 37\% compared to ResNet, demonstrating superior energy efficiency on edge hardware. Finally, MobileViT-XXS achieves the highest throughput (392 ips), confirming its suitability for real-time applications. Overall, MobileViT-XXS achieves lower power consumption, reduced latency, and higher throughput compared to the baseline, making it an ideal backbone for efficient few-shot learning edge AI deployment.

\section{Conclusion}

In this paper, we propose a method for few-shot learning on Edge AI devices using the MobileViT backbone enhanced with knowledge distillation. This approach allows a lightweight student model to inherit the generalization ability of a larger teacher, achieving strong 1-shot and 5-shot classification performance while remaining compact and efficient.
Experiments on MiniImageNet and additional benchmark datasets demonstrate significant improvements over ResNet12 with EASY, reducing parameters by 69\% and FLOPs by 88\% while increasing accuracy by up to 14\% for 1-shot tasks. Deployment on a Jetson Orin Nano confirmed real-world applicability, with MobileViT-XXS lowering dynamic energy consumption by 37\% and achieving a latency of 2.6 ms per inference compared to 3.6 ms for ResNet12.
These results show that combining hybrid CNN-Transformer architectures with knowledge distillation can yield few-shot learning models that are accurate and energy-efficient, making them highly suitable for practical Edge AI applications.

\bibliographystyle{IEEEbib}
\bibliography{biblio}

\end{document}

%% file: colors.tex
\usepackage{xcolor}

\definecolor{darkWhite}{rgb}{0.96,0.96,0.96}
\definecolor{bluekeywords}{rgb}{0.13,0.13,1}
\definecolor{greencomments}{rgb}{0,0.5,0}
\definecolor{redstrings}{rgb}{0.9,0,0}

\definecolor{Comment}{RGB}{97,161,176}

\definecolor{btfGreen}{RGB}{51,160,44}
\definecolor{btfRed}{RGB}{190,60,90}

\definecolor{bleuUni}{RGB}{0, 157, 224}
\definecolor{marronUni}{RGB}{68, 58, 49}

\definecolor{bluecite}{HTML}{009DE0}

\definecolor{Paired_1}{RGB}{31,120,180} 
\definecolor{Paired_2}{RGB}{166,206,227} 
\definecolor{Paired_3}{RGB}{51,160,44} 
\definecolor{Paired_4}{RGB}{178,223,138} 
\definecolor{Paired_5}{RGB}{227,26,28} 
\definecolor{Paired_6}{RGB}{251,154,153} 
\definecolor{Paired_7}{RGB}{255,127,0} 
\definecolor{Paired_8}{RGB}{253,191,111} 
\definecolor{Paired_9}{RGB}{106,61,154} 
\definecolor{Paired_10}{RGB}{202,178,214} 
\definecolor{Paired_11}{RGB}{177,89,40} 
\definecolor{Paired_12}{RGB}{255,255,153} 
\definecolor{Accent_1}{RGB}{127,201,127} 
\definecolor{Accent_2}{RGB}{190,174,212} 
\definecolor{Accent_3}{RGB}{253,192,134} 
\definecolor{Accent_4}{RGB}{255,255,153} 
\definecolor{Accent_5}{RGB}{56,108,176} 
\definecolor{Accent_6}{RGB}{240,2,127} 
\definecolor{Accent_7}{RGB}{191,91,23} 
\definecolor{Accent_8}{RGB}{102,102,102} 
\definecolor{Spectral_1}{RGB}{158,1,66} 
\definecolor{Spectral_2}{RGB}{213,62,79} 
\definecolor{Spectral_3}{RGB}{244,109,67} 
\definecolor{Spectral_4}{RGB}{253,174,97} 
\definecolor{Spectral_5}{RGB}{254,224,139} 
\definecolor{Spectral_6}{RGB}{255,255,191} 
\definecolor{Spectral_7}{RGB}{230,245,152} 
\definecolor{Spectral_8}{RGB}{171,221,164} 
\definecolor{Spectral_9}{RGB}{102,194,165} 
\definecolor{Spectral_10}{RGB}{50,136,189} 
\definecolor{Spectral_11}{RGB}{94,79,162} 
\definecolor{Set1_1}{RGB}{228,26,28} 
\definecolor{Set1_2}{RGB}{55,126,184} 
\definecolor{Set1_3}{RGB}{77,175,74} 
\definecolor{Set1_4}{RGB}{152,78,163} 
\definecolor{Set1_5}{RGB}{255,127,0} 
\definecolor{Set1_6}{RGB}{255,255,51} 
\definecolor{Set1_7}{RGB}{166,86,40} 
\definecolor{Set1_8}{RGB}{247,129,191} 
\definecolor{Set1_9}{RGB}{153,153,153} 
\definecolor{Set1_10}{RGB}{0,0,0} 
\definecolor{Set2_1}{RGB}{102,194,165} 
\definecolor{Set2_2}{RGB}{252,141,98} 
\definecolor{Set2_3}{RGB}{141,160,203} 
\definecolor{Set2_4}{RGB}{231,138,195} 
\definecolor{Set2_5}{RGB}{166,216,84} 
\definecolor{Set2_6}{RGB}{255,217,47} 
\definecolor{Set2_7}{RGB}{229,196,148} 
\definecolor{Set2_8}{RGB}{179,179,179} 
\definecolor{Dark2_1}{RGB}{27,158,119} 
\definecolor{Dark2_2}{RGB}{217,95,2} 
\definecolor{Dark2_3}{RGB}{117,112,179} 
\definecolor{Dark2_4}{RGB}{231,41,138} 
\definecolor{Dark2_5}{RGB}{102,166,30} 
\definecolor{Dark2_6}{RGB}{230,171,2} 
\definecolor{Dark2_7}{RGB}{166,118,29} 
\definecolor{Dark2_8}{RGB}{102,102,102} 
\definecolor{Reds_1}{RGB}{255,245,240} 
\definecolor{Reds_2}{RGB}{254,224,210} 
\definecolor{Reds_3}{RGB}{252,187,161} 
\definecolor{Reds_4}{RGB}{252,146,114} 
\definecolor{Reds_5}{RGB}{251,106,74} 
\definecolor{Reds_6}{RGB}{239,59,44} 
\definecolor{Reds_7}{RGB}{203,24,29} 
\definecolor{Reds_8}{RGB}{165,15,21} 
\definecolor{Reds_9}{RGB}{103,0,13} 
\definecolor{Greens_1}{RGB}{247,252,245} 
\definecolor{Greens_2}{RGB}{229,245,224} 
\definecolor{Greens_3}{RGB}{199,233,192} 
\definecolor{Greens_4}{RGB}{161,217,155} 
\definecolor{Greens_5}{RGB}{116,196,118} 
\definecolor{Greens_6}{RGB}{65,171,93} 
\definecolor{Greens_7}{RGB}{35,139,69} 
\definecolor{Greens_8}{RGB}{0,109,44} 
\definecolor{Greens_9}{RGB}{0,68,27} 
\definecolor{Blues_1}{RGB}{247,251,255} 
\definecolor{Blues_2}{RGB}{222,235,247} 
\definecolor{Blues_3}{RGB}{198,219,239} 
\definecolor{Blues_4}{RGB}{158,202,225} 
\definecolor{Blues_5}{RGB}{107,174,214} 
\definecolor{Blues_6}{RGB}{66,146,198} 
\definecolor{Blues_7}{RGB}{33,113,181} 
\definecolor{Blues_8}{RGB}{8,81,156} 
\definecolor{Blues_9}{RGB}{8,48,107} 

%% file: figure/backbones.tex
\begin{tikzpicture}
    \pgfplotsset{
        width=\linewidth,
        height=0.7\linewidth, 
    }
    \begin{axis}[
        xlabel={Number of Parameters (M)},
        ylabel={Accuracy (\%)},
        xmin=0, xmax=13,
        ymin=55, ymax=75,
        xtick={2,4,6,8,10,12},
        ytick={56,58,60,62,64,66,68,70, 72, 74, 76},
        legend pos=north east,
        ymajorgrids=false,
        grid style=dashed,
    ]
    \addplot[
        color=Paired_5,
        only marks,
        mark=*,
        mark size=3pt
    ]
    coordinates {
        (5.3, 62.03)
        (7.8, 61.20)
        (9.1, 62.11)
        (12.2, 64.55)
    };
    \addlegendentry{EfficientNet}
    \node[above] at (axis cs:5.3, 62.5) {B0};
    \node[above] at (axis cs:7.8, 61.70) {B1};
    \node[above] at (axis cs:9.1, 62.61) {B2};
    \node[above] at (axis cs:12.2, 65.05) {B3};

    \addplot[
        color=Paired_1,
        only marks,
        mark=diamond*,
        mark size=4pt
    ]
    coordinates {
        (2.5, 55.79)
        (5.6, 58.58)
    };
    \addlegendentry{MobileNetV3}
    \node[above] at (axis cs:2.5, 56.09) {S};
    \node[above] at (axis cs:5.6, 58.88) {L};

    \addplot[
        color=Paired_3,
        only marks,
        mark=triangle*,
        mark size=5pt
    ]
    coordinates {
        (0.96, 64.59)
        (1.9, 68.92)
        (4.9, 70.82)
    };
    \addlegendentry{MobileViT}
    \node[above] at (axis cs:0.96, 65.09) {XXS};
    \node[above] at (axis cs:1.9, 69.42) {XS};
    \node[above] at (axis cs:4.9, 71.32) {S};
    \end{axis}
\end{tikzpicture}